\documentclass{article}
\usepackage{spconf,amsmath,amssymb,graphicx,hyperref}

\newcommand{\bX}{\mathbf{X}}
\newcommand{\bz}{\mathbf{z}}
\newcommand{\bh}{\mathbf{h}}

\newcommand{\cL}{\mathcal{L}}
\newcommand{\cD}{\mathcal{D}}

\newcommand{\softmax}{\operatorname{softmax}}

\title{FreqSpaNet: Frequency and Spatial Learning of SFPF for Physical Layer Hardware Integrity Detection}

\name{Xiaoxuan Huang, Jinlong Xu, YiZhe Wang, Meng Zhang, Xian Li, Yuying Bian}
\address{}
\begin{document}
%
\maketitle

\begin{abstract}
Unauthorized hardware replacement can preserve a wireless device's logical identity while altering its physical implementation, posing a challenge to hardware integrity verification. Spatio-frequency polarization fingerprints (SFPFs) capture device-dependent
responses across multiple frequencies and directions, but their
frequency and spatial dimensions exhibit different structural dependencies.
We propose FreqSpaNet, an SFPF representation learning network for
open set hardware anomaly detection. A frequency branch captures local
variations among neighboring frequencies, while a geometry-aware spatial
branch models directional relationships using angular information. The two
representations are combined through adaptive fusion, and complementary pretraining further captures shared information while preserving the distinct characteristics of the frequency and spatial representations.
Experiments show that FreqSpaNet achieves a mean AUROC of
96.31\%, 9.05 points above the baseline. Results
under seven hardware replacement scenarios further
verify the effectiveness of FreqSpaNet.
\end{abstract}
\section{Introduction}
\label{sec:intro}

Unauthorized replacement of a wireless device's hardware module may preserve its communication functions and logical identity while altering its physical implementation, creating a hardware integrity threat. Such hardware changes perturb the device-dependent radiation characteristics of the transmitted signal, which in turn modify its polarization response \cite{xu2022polarization}. Polarization fingerprint (PF) characterizes this response through the complex relation between two orthogonally polarized received components\cite{xu2022polarization,xu2023polarization,xu2022specific}. SFPF extends PF by organizing these responses over multiple frequencies and directions, yielding a structured frequency–spatial representation. Neighboring frequencies tend to show locally correlated responses, whereas different directions capture complementary spatial information. Moreover, the effect of a hardware change is generally nonuniform across frequencies and directions.

Learning an effective representation from SFPF nevertheless presents three
challenges. First, its two dimensions exhibit different structural
dependencies. Second, hardware-induced changes are often
nonuniform across frequencies and directions, so informative
local responses may be weakened by early global aggregation. Third, the
relationships among directions are determined by their physical angular separation rather than by image grid or sequence positions. Generic CNNs and
vision Transformers apply largely homogeneous operators or flatten the
frequency--spatial structure \cite{resnet,vision_transformer}, making them
poorly matched to these SFPF-specific characteristics.

Motivated by these SFPF characteristics, we propose FreqSpaNet, a SFPF representation learning
framework that jointly exploits frequency and spatial information while
retaining the complete angle--frequency structure. Its dual-branch encoder
uses local convolution to capture variations across frequencies and
angular-aware attention to model dependencies among directions.
The two branch representations are then adaptively fused for each input
sample. FreqSpaNet is further trained through frequency--spatial complementary
pretraining, which infers masked SFPF responses while learning both the
information shared by the two branches and the characteristics specific to
each branch. The resulting classifier outputs are calibrated and fused to
detect hardware changes. Experiments show that FreqSpaNet outperforms the evaluated open set systems
built on CNN, Transformer, and MAE backbones
\cite{resnet,vision_transformer,he2022masked}, and remains robust under seven
hardware-replacement scenarios.

\section{Proposed FreqSpaNet}
\label{sec:method}

\subsection{Network Overview}
\label{ssec:overview}

\begin{figure*}[t]
  \centering
  \includegraphics[width=1\textwidth]{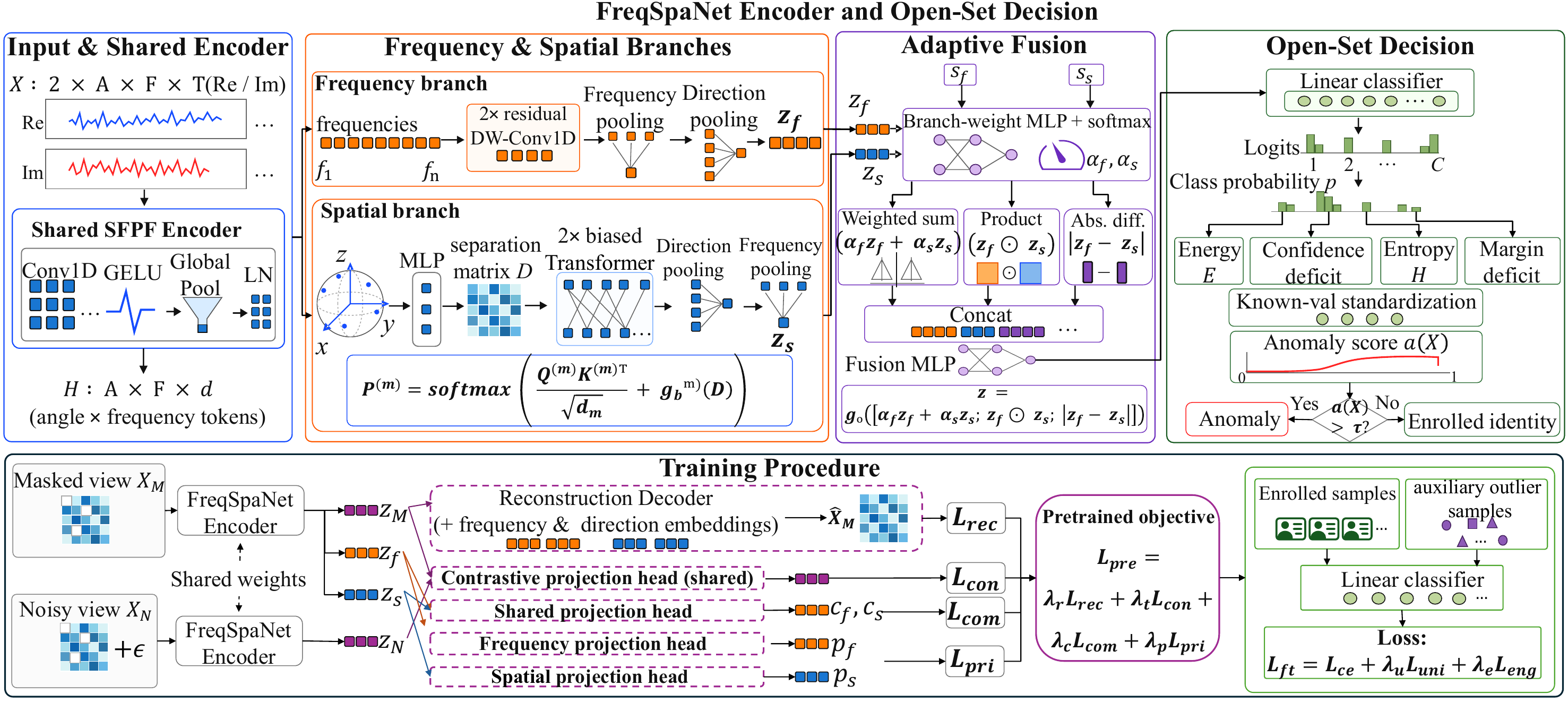}
  \caption{Architecture and learning pipeline of FreqSpaNet.}
  \label{fig:freqspanet_architecture}
\end{figure*}

Figure~\ref{fig:freqspanet_architecture} illustrates the architecture of FreqSpaNet. An input SFPF is denoted by
$\bX\in\mathbb{R}^{2\times A\times F\times T}$, which organizes the complex polarization responses measured over $A$ directions and $F$ frequencies. The first dimension
contains the real and imaginary components, and $T$ denotes the
number of sampling points. We use $A=301$, $F=9$, and $T=1024$. The direction set
is $\cD=\{(\theta_a,\phi_a)\}_{a=1}^{A}$, where $\theta_a$ and $\phi_a$ are
the elevation and azimuth of direction $a$.

For each direction--frequency pair, a shared SFPF encoder maps
$\mathbf{x}_{a,f}\in\mathbb{R}^{2\times T}$ to a token
$\bh_{a,f}\in\mathbb{R}^{d}$. The resulting tensor
$\mathbf{H}\in\mathbb{R}^{A\times F\times d}$ is processed by two parallel
branches. The frequency branch captures local variations across neighboring
frequencies, while the spatial branch models dependencies among
directions. Their outputs, $\bz_f$ and $\bz_s$, are adaptively fused into the
SFPF representation $\bz$, which is mapped to the logits
$\mathbf{o}\in\mathbb{R}^{C}$ of the $C$ enrolled devices. FreqSpaNet is first
pretrained through frequency--spatial complementary learning and then
fine-tuned for enrolled-device classification. At inference, four scores
derived from the classifier outputs are calibrated and fused to detect hardware changes.

\subsection{Frequency--Spatial Encoding and Adaptive Fusion}
\label{ssec:encoding_fusion}

The shared SFPF encoder applies two 1D convolutions,
GELU activations, global pooling, and layer normalization independently to
each $\mathbf{x}_{a,f}$. This produces one token for every measured
direction--frequency response before branch-specific modeling.

The frequency branch adds a frequency embedding to each token and
applies two residual depthwise 1D convolution blocks
\cite{xception}. Their local kernels capture response variations across
neighboring frequencies. Attention pooling first aggregates the frequency
tokens within each direction and then combines all directions to produce
$\bz_f$.

The spatial branch operates on the direction tokens at each
frequency. Direction $(\theta_a,\phi_a)$ is represented by
$\mathbf{u}_a=[\sin\theta_a\cos\phi_a,\sin\theta_a\sin\phi_a,
\cos\theta_a]^{\mathsf T}$, together with the sine and cosine values of its two
angles. A MLP maps this coordinate descriptor to a direction
embedding. Let $D_{ij}=\arccos(\mathbf{u}_i^{\mathsf T}\mathbf{u}_j)$ denote
the angular separation between directions $i$ and $j$. The attention
weights of head $m$ are
\begin{equation}
 \mathbf{P}^{(m)}=\softmax\!\left(
 \frac{\mathbf{Q}^{(m)}\mathbf{K}^{(m)\mathsf T}}{\sqrt{d_m}}
 +g_b^{(m)}(\mathbf{D})\right),
 \label{eq:angular_attention}
\end{equation}
where $\mathbf{Q}^{(m)}$ and $\mathbf{K}^{(m)}$ are the query and key
matrices, $d_m$ is the feature dimension of head $m$, and $g_b^{(m)}$ maps
the angular-separation matrix $\mathbf{D}$ to attention biases. Thus, the spatial branch models interactions between directions using both their learned response features and their angular separation. Pooling over directions and frequencies gives
$\bz_s$.

The relative contribution of the two branches can vary among samples. We
therefore derive two variation statistics from the response amplitude
$R_{a,f,t}=(X_{\mathrm{Re},a,f,t}^{2}
+X_{\mathrm{Im},a,f,t}^{2})^{1/2}$. The frequency variation statistic $s_f$ is obtained by averaging
$|R_{a,f+1,t}-R_{a,f,t}|$ over all pairs of adjacent frequencies, 
directions, and sampling points. For the spatial statistic, we construct an angular neighborhood graph
$\mathcal{G}=(\mathcal{V},\mathcal{E})$ by treating each measured direction as
a vertex and connecting neighboring directions on the acquisition grid. Each edge $(i,j)$ is weighted by
$w_{ij}=\exp(-D_{ij}/\gamma)$, where $\gamma$ controls the decrease with
angular separation. The statistic $s_s$ is the normalized weighted mean of
$|R_{i,f,t}-R_{j,f,t}|$ over graph edges, frequencies, and sampling points.

A fusion network takes $\bz_f$, $\bz_s$, and the normalized pair $(s_f,s_s)$
as input and produces weights $\alpha_f$ and $\alpha_s$ satisfying
$\alpha_f+\alpha_s=1$. The final representation is
\begin{equation}
 \bz=g_o\!\left([\alpha_f\bz_f+\alpha_s\bz_s;\
 \bz_f\odot\bz_s;\ |\bz_f-\bz_s|]\right),
 \label{eq:adaptive_fusion}
\end{equation}
where $g_o$ is the fusion multilayer perceptron. The three
components retain the branch contribution, cross-branch
agreement, and complementary information, respectively.

\subsection{Pretraining and Open Set Detection}
\label{ssec:learning_inference}

During complementary pretraining, a masked view $\bX_M$ and a noisy view
$\bX_N$ of each SFPF are separately processed by the FreqSpaNet encoder with
shared parameters. The FreqSpaNet encoder comprises the shared SFPF encoder,
the frequency and spatial branches, and the adaptive fusion module, producing
the fused representations $\bz_M$ and $\bz_N$ for the two views. A
reconstruction decoder combines $\bz_M$ with the frequency and direction
embeddings to recover the masked responses
\cite{he2022masked,refCMAE}. The reconstruction error on the masked responses
defines $\cL_{\rm rec}$. Meanwhile, a shared projection head maps $\bz_M$ and
$\bz_N$, and the resulting representations are optimized using the supervised
contrastive loss $\cL_{\rm con}$ \cite{khosla2020supervised}.

For the masked view, one shared projection head maps $\bz_f$ and $\bz_s$ to
$\mathbf{c}_f$ and $\mathbf{c}_s$, while two separate projection heads map
them to $\mathbf{p}_f$ and $\mathbf{p}_s$. The former pair represents
information common to both branches, whereas the latter pair represents
information unique to the frequency and spatial branches. We define
$\cL_{\rm com}=1-\cos(\mathbf{c}_f,\mathbf{c}_s)$ to align their common
information and
$\cL_{\rm pri}=|\cos(\mathbf{p}_f,\mathbf{p}_s)|$ to reduce redundancy
between their unique information. The pretraining objective is
\begin{equation}
 \cL_{\rm pre}=\lambda_r\cL_{\rm rec}
 +\lambda_t\cL_{\rm con}
 +\lambda_c\cL_{\rm com}
 +\lambda_p\cL_{\rm pri},
 \label{eq:pretraining}
\end{equation}
where $\lambda_r$,
$\lambda_t$, $\lambda_c$, and $\lambda_p$ weight the four loss terms.

After pretraining, the decoder and projection heads are removed. The pretrained
FreqSpaNet encoder is then jointly fine-tuned with a linear enrolled
classifier, and the parameters of both are updated. During fine tuning, SFPFs collected from devices of different brands and
models from the enrolled devices are used for outlier exposure
\cite{hendrycks2019outlier,liu2020energy}. No hardware-replacement samples
are used during training. The fine-tuning objective is
$\cL_{\rm ft}=\cL_{\rm ce}+\lambda_u\cL_{\rm uni}
+\lambda_e\cL_{\rm eng}$, where $\cL_{\rm ce}$ is the classification loss,
$\cL_{\rm uni}$ encourages uniform predictions for auxiliary outliers, and
$\cL_{\rm eng}$ separates enrolled and auxiliary samples.

For open set detection, let $p_c=\softmax(\mathbf{o})_c$ be the predicted
probability of class $c$. Four anomaly scores are extracted from
$\mathbf{o}$: the energy
$E=-T_e\log\sum_{c=1}^{C}\exp(o_c/T_e)$, maximum-probability uncertainty
$1-\max_c p_c$, normalized entropy
$H=-\sum_{c=1}^{C}p_c\log p_c/\log C$, and probability-margin uncertainty
$1-(p_{(1)}-p_{(2)})$. Here, $T_e$ is the energy temperature and
$p_{(1)}\geq p_{(2)}$ are the two largest class probabilities. The resulting
score vector $\mathbf{r}(\bX)$ is standardized using known-device validation
statistics, giving $\widehat{\mathbf{r}}(\bX)$, and fused as
$a(\bX)=\sigma(\mathbf{w}^{\mathsf T}\widehat{\mathbf{r}}(\bX)+b)$.
The fusion parameters $\mathbf{w}$ and $b$ are learned from known validation
samples and auxiliary outliers without access to the final hardware
anomalies. The threshold $\tau$ is defined by a 5\% false-alarm rate on the known-device
validation set. A sample with
$a(\bX)>\tau$ is reported as a hardware anomaly; otherwise, it is assigned
to the enrolled class with the largest probability.

\section{Experiments}
\label{sec:experiments}

\subsection{Experimental setup}
\label{ssec:setup}

The experiment uses ten wireless devices enrolled in their original hardware configurations. A USRP X310 receiver and an orthogonal dual-polarized antenna are placed 3~m from the device under test, while a motorized pan--tilt platform rotates the device and the receiver remains fixed. Measurements span 913--917~MHz in 0.5 MHz steps, and 1024 complex sampling points are collected for each angle--frequency pair. We use $\theta\in[0^\circ,60^\circ]$ and $\phi\in[0^\circ,120^\circ]$, both with $5^\circ$ steps, yielding 301 physical directions and nine frequency points per SFPF. Five enrolled devices are selected as attack targets, each paired with non-reused replacement hardware and evaluated under all combinations of antenna (A), digital/baseband (D), and RF-front-end (R) replacement, including A, D, R, A+D, A+R, D+R, and A+D+R. For each enrolled device, 600 normal SFPFs at 20~dB are used for training and 100 additional samples are reserved for validation. Auxiliary SFPFs comprise 241 samples from each of three devices whose brands and models are disjoint from those of the enrolled devices and final-test hardware, and are used only for outlier exposure and learning the logistic score fusion weights. Hardware replacement samples are reserved exclusively for final evaluation. At each SNR from 0 to 20~dB in 1 dB increments, the test set contains 241 normal samples per enrolled device and 241 anomalous samples per target device under each hardware replacement scenario.

\subsection{Comparison with generic open set systems}
We compare FreqSpaNet with open set systems built on
ResNet-18 \cite{resnet}, ViT-B/16 and ViT-Tiny/16
\cite{vision_transformer}, and MAE-B/16 \cite{he2022masked}. The evaluated
combinations include maximum softmax probability (MSP)
\cite{hendrycks2017baseline}, energy scoring \cite{liu2020energy}, OpenMax
\cite{bendale2016openmax}, outlier exposure (OE)
\cite{hendrycks2019outlier}, and Deep-SVDD \cite{ruff2018deepsvdd}.

\begin{figure*}[t]
    \centering
    \begin{minipage}{0.42\textwidth}
        \centering
        \includegraphics[width=\linewidth]{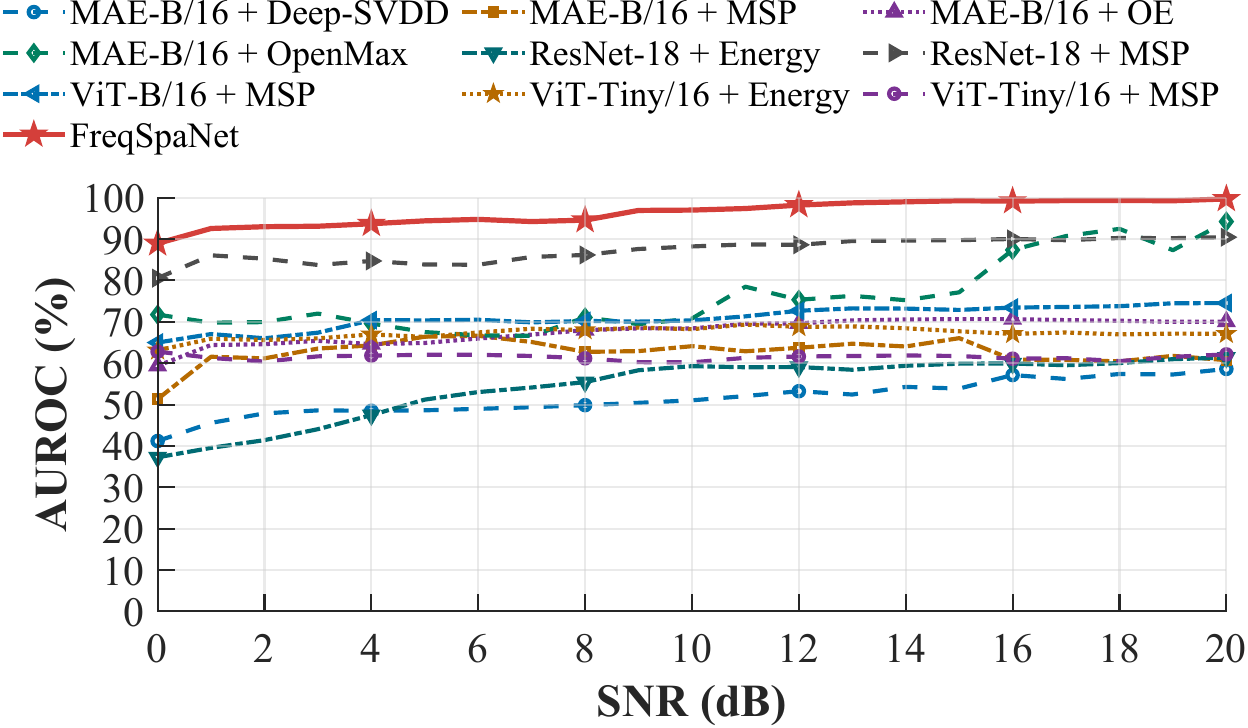}\\[-1mm]
        \textnormal{(a)}
    \end{minipage}
    \hfill
    \begin{minipage}{0.42\textwidth}
        \centering
        \includegraphics[width=\linewidth]{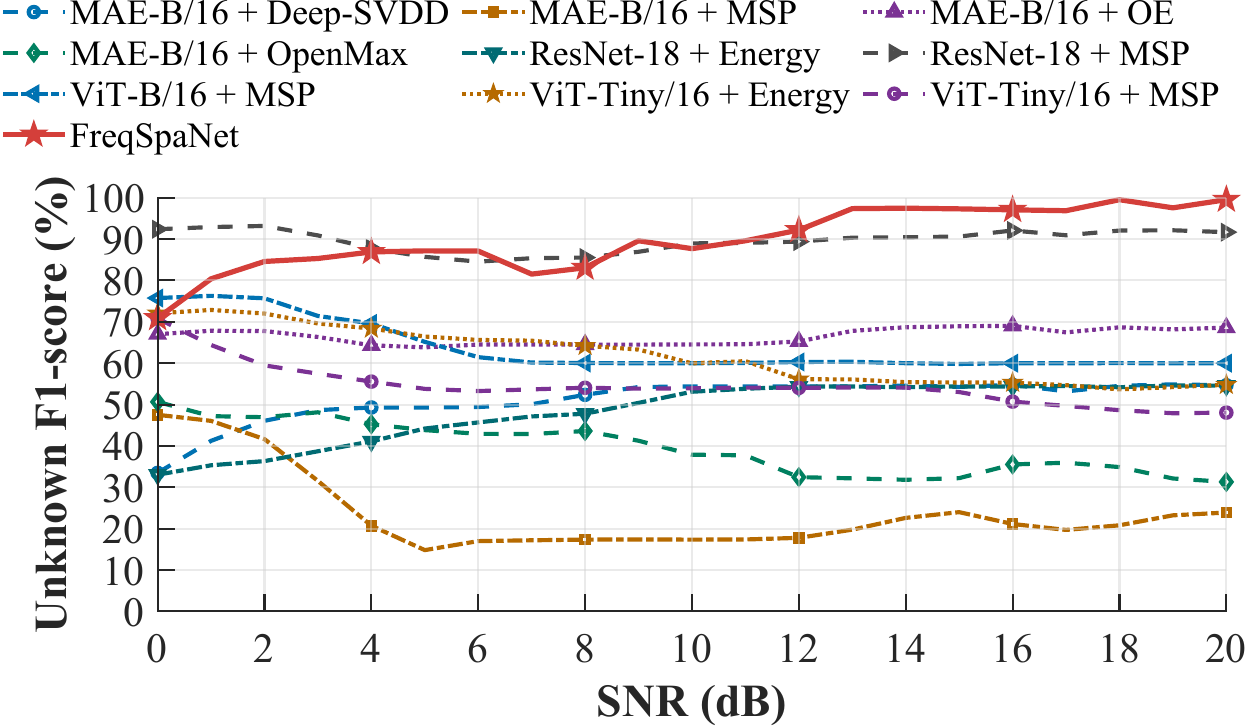}\\[-1mm]
        \textnormal{(b)}
    \end{minipage}
    \caption{Comparison with representative generic open set systems over
    0--20~dB. (a) AUROC; (b) unknown-class F1-score.}
    \label{fig:network_comparison}
\end{figure*}

\begin{figure*}[t]
    \centering
    \begin{minipage}{0.31\textwidth}
        \centering
        \includegraphics[width=\linewidth]{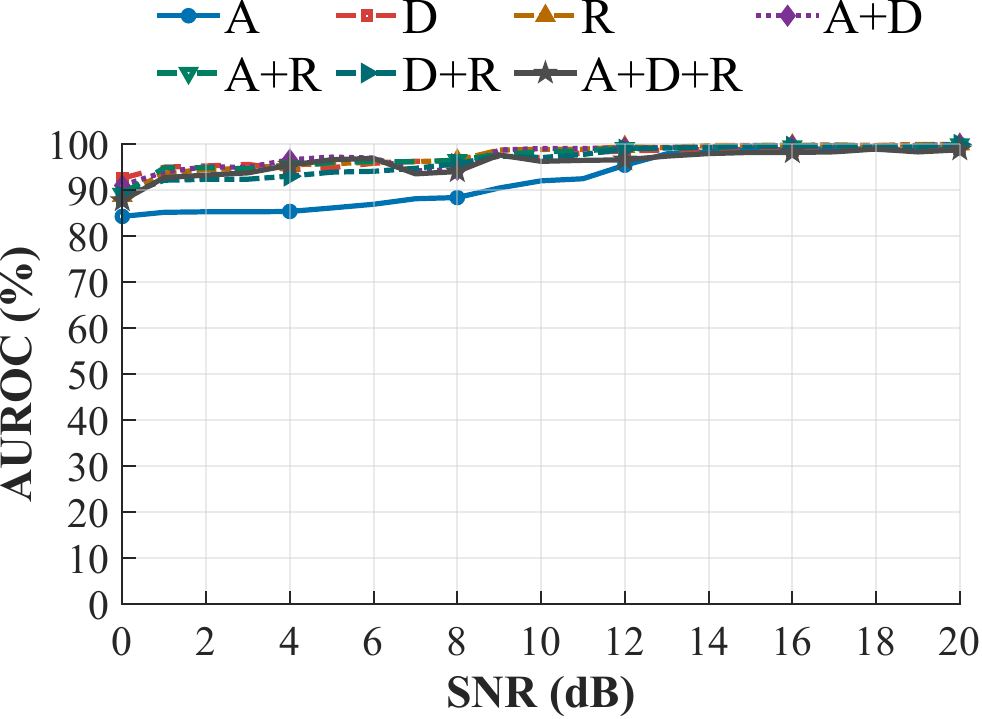}\\[-1mm]
        \textnormal{(a)}
    \end{minipage}
    \hfill
    \begin{minipage}{0.31\textwidth}
        \centering
        \includegraphics[width=\linewidth]{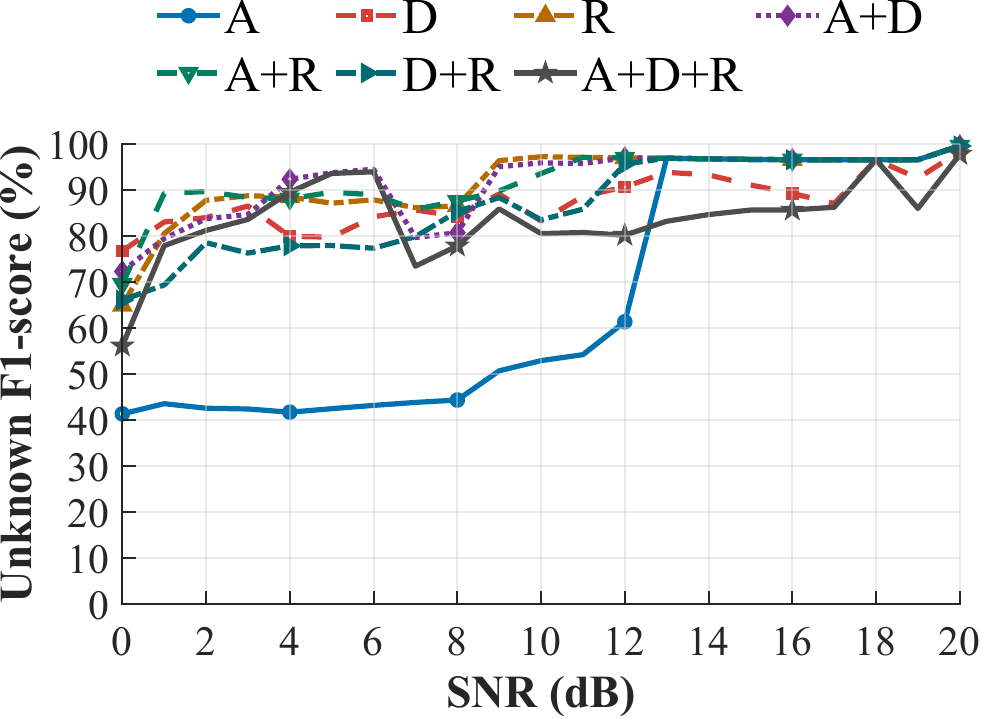}\\[-1mm]
        \textnormal{(b)}
    \end{minipage}
    \hfill
    \begin{minipage}{0.31\textwidth}
        \centering
        \includegraphics[width=\linewidth]{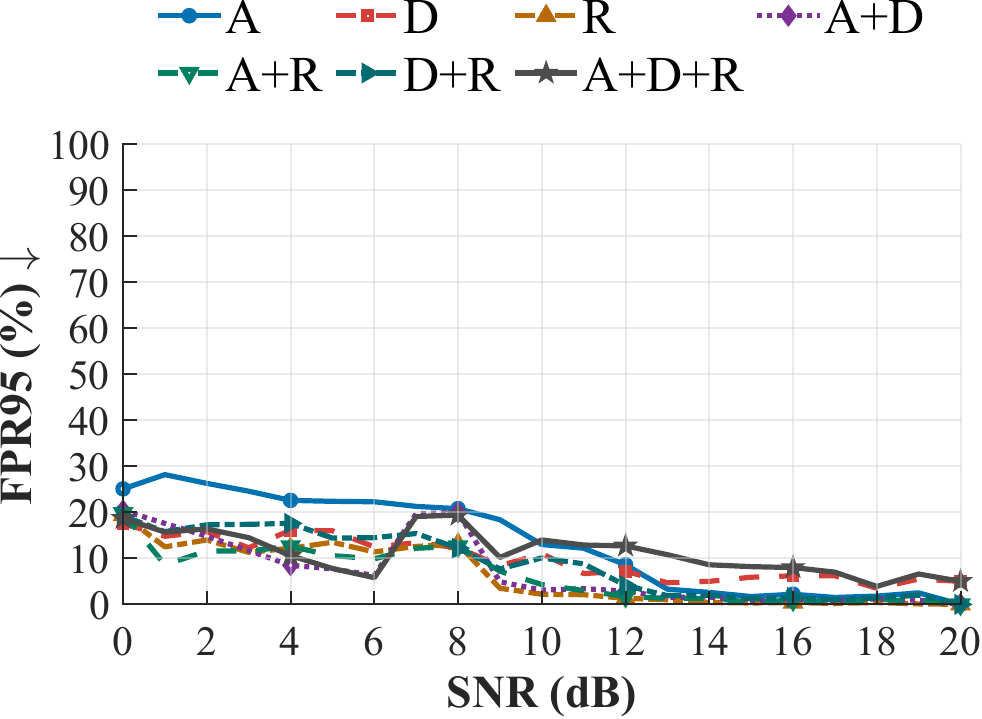}\\[-1mm]
        \textnormal{(c)}
    \end{minipage}
    \caption{FreqSpaNet under seven hardware-replacement scenarios. (a) AUROC; (b) unknown-class F1-score; (c) FPR95.}
    \label{fig:scenario_results}
\end{figure*}

Figure~\ref{fig:network_comparison} compares FreqSpaNet with representative
generic open set systems.  FreqSpaNet achieves a mean AUROC of 96.31\%, exceeding the best evaluated generic baseline on SFPF, ResNet-18+MSP \(87.26\%\), by 9.05 percentage points.  It also
maintains the highest AUROC at every tested SNR, indicating that separately modeling the dependencies along the frequency and spatial dimensions provides more reliable anomaly ranking than generic backbones. FreqSpaNet achieves a mean unknown class F1 score of 89.93\% over the full 0--20~dB range and consistently outperforms the compared methods from 13 to 20~dB.
These results support the use of SFPF-specific frequency and spatial
modeling rather than treating SFPF as a generic image or token sequence. 

\subsection{Robustness to hardware-replacement scenarios}
\label{ssec:scenario_results}

As shown in Fig.~\ref{fig:scenario_results}, antenna replacement is the most challenging case: over 0--5~dB, it yields a mean AUROC of 85.30\% and a mean FPR95 of 24.87\%, whereas the other scenarios achieve mean FPR95 values between 12.47\% and 16.95\%. Nevertheless, from 15 to 20~dB, the averages over all seven scenarios reach 99.33\% AUROC, 95.45\% unknown-class F1-score, and 2.32\% FPR95. These results demonstrate that FreqSpaNet can reliably detect both single- and multi-module hardware replacements.

\subsection{Ablation study}
\label{ssec:ablation}

\begin{table}[t]
 \caption{Ablation results averaged over 0--20~dB (\%).}
 \label{tab:ablation}
 \centering
 \small
 \setlength{\tabcolsep}{1.5pt}
 \begin{tabular}{lccc}
  \hline
  Variant & AUROC $\uparrow$ & FPR95 $\downarrow$ & Bal. acc. $\uparrow$ \\
  \hline
  Frequency only & 93.44 & \textbf{14.42} & 81.12 \\
  Spatial only & 94.60 & 14.69 & 79.57 \\
  w/o direction encoding & 92.36 & 31.89 & 84.15 \\
  w/o angular bias & \textbf{95.39} & 21.53 & \textbf{86.79} \\
  Fixed fusion & 93.89 & 26.13 & 85.70 \\
  \hline
  w/o reconstruction & \textbf{94.11} & \textbf{25.38} & \textbf{85.96} \\
  w/o contrastive & 87.59 & 45.61 & 80.09 \\
  w/o common--private & 92.10 & 38.52 & 85.28 \\
  \hline
  FreqSpaNet (full) & \textbf{96.31} & \textbf{12.30} & \textbf{88.54} \\
  \hline
 \end{tabular}
\end{table}

Table~\ref{tab:ablation} shows that both branches are necessary: using either
branch alone reduces balanced accuracy by more than 7 percentage points.
Removing direction encoding, angular separation bias, or adaptive fusion also
degrades the overall performance, confirming the effectiveness of
geometry-aware spatial modeling and adaptive branch fusion.
The pretraining ablations further show that reconstruction, contrastive
learning, and common--private decomposition all contribute to the final
performance.

\section{Conclusion}
\label{sec:conclusion}

We presented FreqSpaNet, an SFPF-specific representation learning network for
open set hardware anomaly detection. By separately modeling local frequency
dependencies and the angular relationships among directions,
FreqSpaNet preserves the distinct characteristics of the frequency and spatial
dimensions and integrates them through sample-adaptive fusion. Complementary
pretraining further improves the learned representation. Experiments over
0--20~dB show that FreqSpaNet achieves 96.31\% mean AUROC and 89.93\% mean
unknown-class F1-score. Results under seven hardware-replacement scenarios and
ablation studies further demonstrate the effectiveness of the proposed
frequency--spatial representation learning framework.

\bibliographystyle{IEEEbib}
\bibliography{strings,refs}

\end{document}